\documentclass{comsoc2010}

\usepackage{amscd,amsmath,graphicx,latexsym}
\newtheorem{theorem}{Theorem}
\newtheorem{proposition}{Proposition}

\newtheorem{example}{Example}
\newtheorem{definition}{Definition}

\usepackage{color}
\newcommand{\red}[1]{\textcolor{black}{#1}}

\newcommand{\proof}{\noindent {\bf Proof:\ \ }}

\usepackage[ruled,vlined]{algorithm2e}
\newcommand\Omit[1]{}

\title{Stable marriage problems with quantitative preferences}

\author{Maria Silvia Pini, Francesca Rossi, K. Brent Venable, and Toby Walsh\\}

\begin{document}

\begin{abstract}
The stable marriage problem is a well-known problem of matching 
men to women so that no man and woman, who are not married to each 
other, both prefer each other. Such a problem has a wide variety of 
practical applications,  ranging from matching resident doctors to 
hospitals, to matching students to schools or more generally to any 
two-sided market. In the classical stable marriage problem, both men 
and women express a strict preference order over the members of the other sex, 
%and thus they express preferences 
in a qualitative way. Here we consider 
stable marriage problems with quantitative preferences: each man (resp., 
woman) %defines a preference ordering over the women (resp., men) 
%by providing a score for each of them. 
provides a score for each woman (resp., man). 
Such problems are more expressive than the classical stable marriage problems. 
Moreover, in some real-life situations it is more natural to express scores 
(to model, for example, profits or costs) rather than a qualitative preference ordering. 
In this context, we define  new notions of stability and optimality, and we provide 
algorithms to find marriages which are stable and/or optimal 
according to these notions. While expressivity greatly increases 
by adopting quantitative preferences, we show that in most cases 
the desired solutions can be found by adapting existing algorithms 
for the classical stable marriage problem.
\end{abstract}

% Note that the category section should be completed after reference to the ACM Computing Classification Scheme available at
% http://www.acm.org/about/class/1998/.

%\category{H.4}{Information Systems Applications}{Miscellaneous}

%A category including the fourth, optional field follows...
%\category{D.2.8}{Software Engineering}{Metrics}[complexity measures, performance measures]

%General terms should be selected from the following 16 terms: Algorithms, Management, Measurement, Documentation, Performance, Design, Economics, Reliability, Experimentation, Security, Human Factors, Standardization, Languages, Theory, Legal Aspects, Verification.

%\terms{...}

%Keywords are your own choice of terms you would like the paper to be indexed by.

%\keywords{...}

\section{Introduction}

The stable marriage problem (SM) \cite{sm-book} is a well-known problem 
of matching the elements of two sets. 
It is called the {\em stable marriage} problem since the standard formulation
is in terms of men and women, and the matching is interpreted in terms of 
a set of marriages. 
Given $n$ men and $n$ women, where each person expresses a 
strict ordering over the members of the opposite sex, the problem is  
to match the men to the women %off 
so that there are no 
two people of opposite sex who would both rather be matched with each other 
than their current partners. 
If there are no such people, all the marriages are said to be {\em stable}. %"stable".
In \cite{gs} Gale and Shapley proved that 
%for any equal number of men and women, 
it is always possible to 
find a matching that makes all marriages stable, and provided a polynomial %quadratic
time algorithm which can be used to find
one of two extreme stable marriages,
the so-called {\em male-optimal} or {\em female-optimal} solutions. 
%The Gale-Shapley algorithm involves a number of "rounds" 
%(or "iterations") where each UN-engaged man "proposes" 
%to the most-preferred woman to whom he has not yet proposed. 
%
The Gale-Shapley algorithm 
has been used in many real-life scenarios, such as in 
matching hospitals to resident doctors \cite{roth-H}, 
medical students to hospitals, sailors to ships \cite{liebowitz}, 
primary school students to secondary schools \cite{revisited}, 
as well as in market trading \cite{market}. 
%Variants of the stable marriage 
%problem turn up in many domain\cite{irving-med}s. For 
%example, the US Navy has a web-based multi-agent system
%for assigning sailors to ships \cite{liebowitz}. 

%.............................
In the classical stable marriage problem, both men 
and women express a strict preference order over the members of the other sex in a qualitative way. 
Here we consider stable marriage problems with quantitative preferences. In such problems each man (resp., 
woman) provides a score for each woman (resp., man).  
Stable marriage problems with quantitative preferences are interesting since they are more expressive than the classical stable marriage problems, since in classical stable marriage problem a man (resp., a woman) cannot express how much he (resp., she) prefers a certain woman (resp., man).  
Moreover, they are useful in some real-life situations where it is more natural to express scores, 
that can model notions such as
profit or cost, rather than a qualitative preference ordering.  In this context, we define 
 new notions of stability and optimality, we compare such notions with the classical ones,  
 and we show algorithms to find marriages which are stable and/or optimal 
according to these notions. 
While expressivity increases 
by adopting quantitative preferences, we show that in most cases 
the desired solutions can be found by adapting existing algorithms 
for the classical stable marriage problem.

%\section{Related work}
Stable marriage problems with quantitative preferences have been studied also in  \cite{Gus87,irleagus}. 
However,  they solve these problems by looking at the 
stable marriages that maximize the sum of the weights of the married pairs, where the weights depend
on the specific criteria used to find an optimal solution, that can be minimum regret criterion \cite{Gus87}, 
the egalitarian criterion \cite{irleagus} or the Lex criteria \cite{irleagus}. 
Therefore, they consider as stable the same marriages that are stable when we don't consider the weights. 
We instead use the weights to define new notions of stability that may lead to stable marriages that are different from the classical case. They may rely on the difference of weights that 
a person gives to two different people of the other sex, 
or by the strength of the link of the pairs (man,woman), i.e., how much a person of the pair wants to be married with the other person of the pair. 
%\item From Marriages to Coalitions: A Soft CSP Approach 
The classical definition of stability for stable marriage problems with quantitative preferences has been 
considered also in \cite{bista-stable} that has used a semiring-based soft constraint approach \cite{jacm} 
to model and solve these problems.   
%We instead define  more sofisticated notions of stability and we present ad-hoc algorithms that look for every new notion.  
%Also in \cite{bista-stable} stable marriage problems with quantitative preferences have been investigated. 
%They have modelled such problems as soft constraint problems and they have used as optimization criterion a criterion 
%that depends on the chosen semiring (such as Weighted and Fuzzy) thus solving with a unique  solver 
%also different problems already solved in the literature such as minimizing the egalitarian cost. 
%It relates the optimal stable marriage problem and soft constraint satisfaction as done in \cite{gent}  
%for the classical stable marriage problem. The optimization criterion consider in \cite{bista-stable} 
%depends on the chosen semiring (such as Weighted and Fuzzy) which can be used to solve problems already proposed in the literature such as minimizing the egalitarian cost. 
%While in \cite{bista-stable} they have considered the classical definition of stability, 
%we define  more sofisticated notions of stability and we present 
%ad-hoc algorithms that look for every new notion.  

The paper is organized as follows. 
In Section \ref{back} we give the basic notions of classical stable marriage problems, stable marriage problems with partially ordered preferences and stable marriage problems with quantitative preferences (SMQs). 
In Section \ref{sec-alpha} we introduce a new notion of stability, called $\alpha$-stability for SMQs, which depends on the difference of scores that every person gives to two different people of the other sex, and we compare it with the classical notion of stability. 
Moreover, we give a new notion of optimality, called lex-optimality, 
to discriminate among the new stable marriages, which depends on a voting rule. We show that there is a unique optimal stable marriage and we give an algorithm to find it.  
In Section \ref{sec-link} we introduce other notions of stability for SMQs that are based on  the strength of the link of the pairs (man,woman), we compare them with the classical stability notion, and we show how to find marriages that are stable according to these notions with the  highest global link. 
In Section \ref{sec-concl} we summarize the results contained in this paper, and we give some hints for future work.

\section{Background}
\label{back}

We now give some basic notions on classical stable marriage problems, 
stable marriage problems with partial orders, and stable marriage problems with quantitative preferences. 

\subsection{Stable marriage problems}
%\label{sm}

A {\em stable marriage problem} (SM) \cite{sm-book} of size $n$ 
is the problem of finding a stable marriage between $n$ men and $n$ women.
Such men and women each have a preference ordering over the members of the other sex.
A marriage is a one-to-one correspondence between \red{ men} and women. 
Given a marriage $M$, a man $m$, and a woman $w$, 
the pair $(m,w)$ is a  {\em blocking pair} for $M$ if $m$ prefers $w$ to his partner in $M$ 
and $w$ prefers $m$ to her partner in $M$. 
A marriage is said to be {\em stable} if it does not contain blocking pairs. 
%there are no two people of opposite sex who would both rather
%be married to each other than their current partners.

The sequence of all preference orderings of men and women is usually called a {\em profile}.
In the case of classical stable marriage problem (SM), a profile is a 
sequence of strict total orders. 

%\begin{definition}[profile] 
%Given $n$ men and $n$ women, a profile is a sequence
%of $2n$ strict total orders (i.e., transitive and complete binary relations), $n$ over the men and $n$ over the women.
%\end{definition}
%
%Each of such total orders is associated to one man or woman, and 
%models his/her preference ordering over the members of the other sex.

%\begin{definition}[stable matching problem (SM)]
%\end{definition}

%Given a profile, the stable marriage problem (SM) \cite{gs}
%is the problem of finding  a matching between men and women.
%The goal is to marry the men to the women such that there 
%are no two people of opposite sex who would both rather 
%be married to each other than their current partners. 
%If there are no two such people, the matching is said to be stable. 
%We will assume that the total number of agents (i.e., men + women) is $N$.

Given a SM $P$, there may be many stable marriages for $P$. However, 
it is interesting to know that there is always at least one stable marriage. 

%\begin{definition}[feasible partner]
Given an SM $P$, a {\em feasible partner} for a man $m$ (resp., a woman $w$) 
is a woman $w$ (resp., a man $m$) such that there is a stable marriage for $P$ 
where $m$ and $w$ are married. 
%\end{definition}

The set of all stable marriages for an SM forms a lattice,
where a stable marriage $M_1$ dominates another stable marriage $M_2$
if men are happier (that is, are married to more or equally preferred women)
in $M_1$ w.r.t. $M_2$. The top 
of this lattice is the stable marriage where men are most satisfied, and it is usually called 
the {\em male-optimal} stable marriage. Conversely, 
the bottom is the stable marriage where men's preferences are least satisfied (and women are happiest, so it is usually called the {\em female-optimal} stable marriage).    
Thus, 
%\begin{definition}[male (resp., female) optimal matching]
%Given an SM $P$, 
a stable marriage is %(resp., female) 
male-optimal iff every man %(resp., woman) 
is paired with his %(resp., her) 
highest ranked feasible partner. % in $P$. 
%\end{definition}

%\subsection{Gale-Shapley algorithm}

The {\em Gale-Shapley} (GS) {\em algorithm}  \cite{gs}
%, shown below as Algorithm \ref{gs-alg}, 
is a well-known algorithm to solve the SM problem.  
At the start of the algorithm, each person is free and becomes engaged during the execution of
the algorithm. Once a woman is engaged, she never becomes free again (although to whom she is
engaged may change), but men can alternate between being free and being engaged. The following
step is iterated until all men are engaged: 
choose a free man $m$, and let $m$ propose to the most preferred woman $w$ on his preference list,
such that $w$ has not already rejected $m$. 
If $w$ is free, then $w$ and $m$ become engaged. If $w$ is engaged
to man $m$', then she rejects the man ($m$ or $m$') that she least prefers, 
and becomes, or remains, engaged to the other man. 
The rejected man becomes, or remains, free.
When all men are engaged, the engaged pairs form the male 
optimal stable matching. 
It is female optimal, of course, if the roles of male and female 
participants in the algorithm were interchanged. 

This algorithm needs a number of steps that, in the worst case, is quadratic in $n$ (that is, the number of men), 
and it guarantees that,  
if the number of men and women coincide, and all participants 
express a strict order over all the members of the other group,
everyone gets married, 
and the returned matching is stable. 
%Notice that, since the input is a profile, 
%then the algorithm is linear in the size of the input.  

%Note that the pairing generated by the Gale-Shapley algorithm is male optimal.   
%
%\begin{example}
%\label{ex0}
%Assume $n=3$. Let $W=\{w_1,w_2,w_3\}$ and  $M=\{m_1,m_2,m_3\}$
%be respectively the set of women and men.
%The following sequence of strict total orders defines a profile:
%\begin{itemize}
%\item $m_1:w_1>w_2>w_3$ (i.e., the man $m_1$ prefers the woman $w_1$ to $w_2$ to $w_3$),
%\item $m_2:w_2>w_1>w_3$,
%\item $m_3:w_3>w_2>w_1$,
%\item $w_1:m_1>m_2>m_3$,
%\item $w_2:m_3>m_1>m_2$,
%\item $w_3:m_2>m_1>m_3$
%\end{itemize}
%For this profile, the Gale-Shapley algorithm returns the male optimal solution
%$\{(m_1,w_1),(m_2,w_2),(m_3,w_3)\}$. On the other hand, the female optimal solution
%is $\{(w_1,m_1),(w_2,m_3),(w_3,$ $m_2)\}$. \hfill $\Box$
%\end{example}

\begin{example}
\label{ex0}
Assume $n=2$. Let $\{w_1,w_2\}$ and  $\{m_1,m_2\}$ 
be respectively the set of women and men. 
The following sequence of strict total orders defines a profile:
%$\{ m_1:w_1>w_2>w_3$ (i.e., man $m_1$ prefers woman $w_1$ to $w_2$ to $w_3$); 
%\mbox{ } $m_2:w_2>w_1>w_3;$ 
%\mbox{ } $m_3:w_3>w_2>w_1\}$ 
%$\{ w_1:m_1>m_2>m_3;$ 
%\mbox{ } $w_2:m_3>m_1>m_2;$
%\mbox { } $w_3:m_2>m_1>m_3\}.$ 
\begin{itemize}
\item $m_1:w_1>w_2$ (i.e., man $m_1$ prefers woman $w_1$ to woman $w_2$),
\item $m_2:w_1>w_2$,
\item $w_1:m_2>m_1$,
\item $w_2:m_1>m_2.$
\end{itemize}
For this profile, the male-optimal solution is 
$\{(m_1,w_2),(m_2,w_1)\}$. For this specific profile the female-optimal 
stable marriage coincides with the male-optimal one.  \hfill $\Box$
\end{example}

%The {\em Extended Gale-Shapely algorithm} \cite{sm-book} is just the GS algorithm \cite{gs} where, 
%whenever the proposal of a man $m$ to a woman $w$ is accepted,
%in $w$'s preference list all men less desirable than $m$ are deleted, and
%$w$ is deleted from the preference lists of all such men. This means that, every time that a woman %receives a proposal from a man, she accepts since only more preferred men can propose to her.  

\subsection{Stable marriage problems with partially ordered preferences}

In SMs, each preference ordering is a strict total order over the members of the other sex. 
More general notions of SMs allow preference orderings to be partial \cite{rev-manlove}. %\cite[irving84}.
This allows for the modelling of both indifference (via ties) 
and incomparability (via absence of ordering) between members of the other sex.
%We assume now that men and women express their preferences via partial orders.
%The notions given in the previous section %\ref{back} 
%can be generalized as follows.
%So, for our study, a profile will be 
%\begin{definition}[partially ordered profile]
%Given $n$ men and $n$ women, a profile is 
In this context, a stable marriage problem  is defined by a sequence of $2n$ partial orders,
%(i.e., reflexive, antysimmetric and transitive binary relations), 
$n$ over the men and $n$ over the women.
%\end{definition}
We will denote with SMP a stable marriage problem with such partially ordered preferences.
 
%\begin{definition}[SMP] 
%Given a partially ordered profile, a stable matching problem with partial orders (SMP) 
%is just a SM where men's preferences and women's preference are partially ordered. 
%\end{definition}

Given an SMP, we will sometimes use the notion of a {\em linearization} 
of such a problem, which is obtained by linearizing the preference orderings 
of the profile in a way that is compatible with the given partial orders.
%considering 
%\begin{definition}[linearization of an SMP]
%A linearization of an SMP is an SM that is obtained by giving a strict ordering 
%to all the pairs that are not strictly ordered such that the resulting ordering is transitive.
%\end{definition}
 
%\begin{definition}[super-stable matching in SMPs]
%\end{definition}

%\begin{definition}[strong-stable matching in SMPs]
%\end{definition}

%\begin{definition}[weakly-stable matching in SMP]
%\label{weak}
A marriage $M$ for an SMP is said to be {\em weakly-stable} 
if it does not contain blocking pairs. Given a man $m$ and a woman $w$, the pair $(m,w)$ 
is a blocking pair if $m$ and $w$ are not married to each other in $M$ and 
each one {\em strictly prefers} the other to his/her current partner. 
%\end{definition}
%\begin{definition}[weakly-stable matching]
%A matching is weakly-stable if there is no pair $(x,y)$ such that each one strictly 
%prefers the other to his/her current partner. 
%\end{definition}
%
%\begin{definition}[strongly-stable matching]
%A matching is strongly-stable if there is no pair $(x,y)$ such that $x$ strictly prefers 
%$y$ to his/her current partner, and $y$ strictly prefers $x$ to his/her current partner or 
%is indifferent between them. 
%%he/she considers them in a tie or incomparable. 
%\end{definition}
%
%\begin{definition}[super-stable matching]
%A matching is super-stable if there is no pair $(x,y)$ such that each of whom either 
%strictly prefers the other to his/her current partner or %is indifferent between them. 
%\end{definition}
%\begin{definition}[feasible partner in SMP]
%Given an SMP $P$, a feasible partner for a man $m$ (resp., woman $w$) is a woman $w$ 
%(resp., man $m$) such that there is a weakly stable marriage for $P$ 
%where $m$ and $w$ are married. 
%\end{definition}

%A weakly stable marriage is said to be {\em undominated} %male-optimal 
%if there is no other weakly stable marriage that dominates it. 
A weakly stable marriage $M$ dominates a weakly stable marriage $M'$ iff 
for every man $m$, $M(m)\geq M'(m)$ and 
there is a man $m'$ s.t. $M(m')>M'(m')$. 
Notice that there may be more than one undominated weakly stable marriage for an SMP.

\subsection{Stable marriage problems with quantitative preferences}

In classical stable marriage problems, men and women express 
only qualitative preferences over the members of the other sex. 
For every pair of women (resp., men), every man (resp., woman) 
states only that he (resp., she) prefers a woman (resp., a man) 
more than \red{another} one.  However, he (resp., she) cannot express how much 
he (resp., she) prefers such a woman (resp., a man). 
This is \red{nonetheless} possible in stable marriage problems with quantitative preferences.

A {\em stable marriage problem with quantitative preferences} (SMQ) \cite{irleagus}  
is a classical SM where every man/woman gives also a 
numerical preference value for every member of the other sex, that represents how much he/she 
prefers such a person. \red{ Such preference values are natural numbers and}
higher preference values denote a more preferred item. 
Given a man $m$ and a woman $w$, the {\em preference value} for  man $m$ (resp., woman $w$) 
of  woman $w$ (resp., man $m$) will be denoted by $p(m,w)$  (resp., $p(w,m)$). 
%It represents how much  $m$ (resp., $w$) wants to be married with $w$ (resp., $m$). 
%\end{definition}

%Given an SMQ $P$ and a marriage $M$, the {\em men's (resp., women's) weight} 
%of $M$ in $P$, denoted by $w_{men}(M,P)$ (resp., $w_{women}(M,P)$), is given 
%by the sum of the weights that each man (resp., woman) gives to his/her partner in $M$. 
%Given an SMQ $P$ and a marriage $M$, the {\em preference weight} of $M$ in $P$, denoted by $w(M,P)$, is given by 
%$w_{men}(M,P)+w_{women}(M,P)$. 

\begin{example}
\label{ex-smw}
Let $\{w_1,w_2\}$ and  $\{m_1,m_2\}$
be respectively the set of women and men. 
%An instance of an SMW is defined by a profile and a weights' vector. 
%Assume the profile is the same shown in \ref{ex0} and that
%the weights' vector is  ${{9,1},{3,2},{2,1},{3,1}}$, that means that 
%$m_1$ prefers with degree $9$ to be married with woman $w_1$ and with degree $1$ to be married to 
%the woman $w_2$, $m_2$ prefers with degree $3$ to be married with woman $w_1$ and with degree $2$ to be
% married to the woman $w_2$, $w_1$ prefers with degree $2$ to be married with the man $m_1$ and with 
% degree $1$ to be married to the man $m_2$, and $w_2$ prefers with degree $3$ to be married with the man
%  $m_1$ and with degree $1$ to be married to the man $m_2$. 
An instance of an SMQ is the following: 
\begin{itemize}
\item \red{$m_1:w_1^{[9]} >w_2^{[1]}$} (i.e.,  man $m_1$ prefers  woman $w_1$ to woman $w_2$, 
                              and he prefers $w_1$ with value $9$ and $w_2$ with value $1$),
\item \red{$m_2:w_1^{[3]} >w_2^{[2]}$,
\item $w_1:m_2^{[2]} > m_1^{[1]}$,
\item $w_2:m_1^{[3]} >m_2^{[1]}.$}
\end{itemize} 
The numbers written into the round brackets identify the preference values. 
%For example, the weight of the woman $w_1$ for the man $m_1$, i.e., $p(m_1,w_1)$, is $9$. 
%The weight of the marriage $\{(m_1,w_1),(m_2,w_2)\}$ is $9+2+1+1=13$. 
%If we forget about the weights, the profile above corresponds to the one shown in Example \ref{ex0}.  
 \hfill $\Box$
\end{example}

%\section{Stability and optimality for SMQs}

In \cite{irleagus} they consider stable marriage problems with quantitative preferences 
by looking at the stable marriage that maximizes the sum of the preference values. 
Therefore, they use the classical definition of stability and they use  preference values 
only when they have to look for the optimal solution. 
We want, instead, %to give more importance to the 
to use preference values also to define new notions of stability and optimality. 

We will introduce new notions of stability and optimality that are based on the quantitative preferences expressed by the agents  and we will show how to find them by adapting the classical Gale-Shapley algorithm \cite{gs} for SMs described in Section \ref{back}.

%A marriage in SMWs is {\em stable} if it does not contain blocking pairs, i.e., 
%if there are not pairs of men and women that prefer to be married each other than 
%with their current partners, or if there is such pairs, 
%but their link strength is lower than the sum of the 
%link strengths of these people with their current partners. More precisely:

%\begin{definition}[blocking pair]
%Consider a marriage $M$, a man $m$ and a woman $w$ that are not married in $M$. 
%Let $m'$ (resp., $w'$) be the parner of $w$ (resp., $m$) in $M$. 
%The pair $(m,w)$ is a blocking pair for $M$ if 
%$m$ prefers $w$ to $w'$, $w$ prefers $m$ to $m'$, and 
%$l(m,w)>l(m,w')+l(m',w)$. 
%$l(m,w)>l(m,w')$ or $l(m,w)>l(m',w)$...................
%\end{definition}

\section{$\alpha$-stability}
\label{sec-alpha}

A simple generalization of the classical notion of stability requires 
that there are not two people that prefer {\em with at least degree} $\alpha$ 
(where $\alpha$ is a natural number) %natural number) 
to be married to each other rather than to their current partners.

\begin{definition}[$\alpha$-stability]
Let us consider a natural number $\alpha$ with $\alpha\geq 1$. 
Given a marriage $M$, a man $m$, and a woman $w$, 
the pair $(m,w)$ is an $\alpha$-blocking pair for $M$ if the following conditions hold: 
\begin{itemize}
\item $m$ prefers $w$ to his partner in $M$, say $w'$, 
\red{ by at least}  $\alpha$ (i.e., $p(m,w)-p(m,w')\geq \alpha$), 
\item $w$ prefers $m$ to her partner in $M$, say $m'$,
 \red{ by at least}  $\alpha$ (i.e., $p(w,m)-p(w,m')\geq \alpha$). 
\end{itemize}
A marriage is $\alpha$-stable if it does not contain $\alpha$-blocking pairs. 
A man $m$ (resp., woman $w$) is $\alpha$-feasible for woman $w$ (resp., man $m$) 
if $m$ is married with $w$ in some $\alpha$-stable marriage. 
\end{definition}

\subsection{Relations with classical stability notions}
  
Given an SMQ $P$, let us denote with $c(P)$, the classical SM problem obtained from $P$ 
by considering only the preference orderings induced by the preference values of $P$.  

\begin{example}
Let us consider the SMQ, $P$, shown in Example \ref{ex-smw}. 
The stable marriage problem $c(P)$ is shown in Example \ref{ex0}. \hfill $\Box$. 
\end{example}

%If we don't consider the preference values given by the agents, i.e., when we associate to a every 
%person in position $i$ in a preference ordering preference value $n-i$, 
%and we assume that 
If $\alpha$ is equal to $1$, then the $\alpha$-stable marriages of $P$ coincide with the stable marriages of $c(P)$. 
However, in general,  $\alpha$-stability  allows us to have more marriages 
that are stable according to this definition, since we have a more \red{relaxed} notion of 
blocking pair. In fact, a pair $(m,w)$ is an $\alpha$-blocking if both $m$ and $w$ prefer each other to their current partner {\red {\em by at least $\alpha$}} and thus pairs $(m',w')$ where $m'$ and $w'$ prefer each other to their current partner  of less than $\alpha$ are not considered $\alpha$-blocking pairs. 
 
\red{The fact that $\alpha$-stability leads to a larger number of stable marriages w.r.t. the classical case is important to allow new stable marriages where some men, for example the most popular ones, may be married with partners better than all the feasible ones according to the classical notion of stability.} 
%as optimal also some marriages that are better, according to specific criteria due to the preference values, 
%than those considered stable in the classical case 
%where preference values are not considered. 

Given an SMQ $P$, let us denote with 
$I_\alpha(P)$ the set of the $\alpha$-stable marriages of $P$ and with 
$I(c(P))$ the set of the stable marriages of $c(P)$. We have the following results.

\begin{proposition}
Given an SMQ $P$, and a natural number $\alpha$ with $\alpha\geq 1$,  
\begin{itemize}
\item if $\alpha=1$,  $I_\alpha(P)=I(c(P))$; 
\item if $\alpha>1$, $I_\alpha(P) \supseteq I(c(P))$. 
\end{itemize}
\end{proposition}

Given an SMP $P$, the set of $\alpha$-stable marriages of $P$ contains  
the set of stable marriages of $c(P)$, since the $\alpha$-blocking pairs of $P$ are a subset 
of the blocking pairs of $c(P)$. 

Let us denote with $\alpha(P)$ the stable marriage with incomparable pairs obtained from an SMQ  
$P$ by setting as incomparable every pair of people that don't differ for at least $\alpha$, and with 
$I_w(\alpha(P))$ the set of the weakly stable marriages of $\alpha(P)$. 
It is possible to show that the set of the weakly stable marriages of $\alpha(P)$ 
coincides with the set of the $\alpha$-stable marriages of $P$. 

\begin{theorem}
\label{teo-pa}
Given an SMQ $P$, $I_\alpha(P) = I_w(\alpha(P))$. 
\end{theorem}

\proof{
We first show that $I_\alpha(P) \subseteq I_w(\alpha(P))$. 
Assume that a marriage $M\not \in I_w(\alpha(P))$, we now show that $M\not \in I_\alpha(P)$. 
If $M\not \in I_w(\alpha(P))$, then there is a pair (man,woman), say $(m,w)$, in $\alpha(P)$
 such that $m$ prefers $w$ to \red{his partner in $M$}, say $w'$, 
 and $w$ prefers $m$ to \red{her partner in $M$}, say $m'$. 
 By definition of $\alpha(P)$, 
 this means that $m$ prefers $w$ to $w'$ \red{ by at least} degree $\alpha$ 
  and $w$ prefers $m$ to $m'$ \red{ by at least} degree $\alpha$ in $P$, and so $M\not \in I_\alpha(P)$. 
Similarly, we can show that $I_\alpha(P) \supseteq I_w(\alpha(P))$. 
In fact, if $M\not \in I_\alpha(P)$, then there is a pair (man,woman), say $(m,w)$, in $P$
 such that $m$ prefers $w$ to $w'$ \red{ by at least} degree $\alpha$ 
 and $w$ prefers $m$ to $m'$ \red{ by at least} degree $\alpha$. 
 By definition of $\alpha(P)$, this means that
  $m$ prefers $w$ to $w'$ and $w$ prefers $m$ to $m'$ in $\alpha(P)$ 
  and so $M\not \in I_w(\alpha(P))$, i.e., 
  $M$ is not a weakly stable marriage for $\alpha(P)$. \hfill $\Box$\\
}

This means that, given an SMQ $P$, every algorithm that is able to find a weakly stable marriage for $\alpha(P)$ 
provides an $\alpha$-stable marriage for $P$.

\begin{example}
\label{exalpha}
Assume that $\alpha$ is $2$. 
Let us consider the following instance of an SMQ, say $P$.  
\begin{itemize}
\item \red{$m_1:w_1^{[3]} > w_2^{[2]}$ 
\item $m_2:w_1^{[4]} > w_2^{[2]}$,
\item $w_1:m_1^{[8]} > m_2^{[5]}$,
\item $w_2:m_1^{[3]} >m_2 ^{[1]}.$}
\end{itemize} 
The SMP $\alpha(P)$ is the following: 
\begin{itemize}
\item $m_1:w_1  \bowtie w_2$ (where $\bowtie$ means incomparable), 
\item $m_2:w_1 > w_2$,
\item $w_1:m_1 > m_2$,
\item $w_2:m_1 >m_2.$
\end{itemize} 
The set of the $\alpha$-stable marriages of $P$, 
that coincides with the set of the weakly stable marriages of $\alpha(P)$, 
by Theorem \ref{teo-pa}, contains the following marriages: 
$M_1=\{(m_1,w_1),(m_2,w_2)\}$ and 
$M_2=\{(m_1,w_2),(m_2,w_1)\}$. 
%Notice that $M_1$ does not dominate $M_2$ and $M_2$ does not dominate $M_1$.
\hfill $\Box$ 
\end{example}

\red{On the other hand, not all stable marriage problems with partially ordered preferences 
can be expressed as stable marriage problems with quantitative preferences 
such that the stable marriages in the two problems coincide.
More precisely, given any SMP problem $P$, we would like to be able to generate a corresponding 
SMQ problem $P'$ and a value $\alpha$ such that, in $P'$,
the weights of elements ordered in $P$ differ more than $\alpha$, while those 
of elements that are incomparable in $P$ differ less than $\alpha$.
Consider for example the case of a partial order over six elements, defined as follows:
$x_1>x_2>x_3>x_4>x_5$ and $x_1>y>x_5$. Then there is no way to choose a value $\alpha$ 
and a linearization of the partial order such that the weights of 
$x_i$ and $x_j$ differ for at least $\alpha$, for any i,j between 1 and 5, 
while at the same time the weight of $y$ and each of the $x_i$'s differ for less than $\alpha$.  
}

%In fact, if we consider an SMP, say $P$, where the preference ranking of a man (resp., woman) is defined by an arbitrarily long chain of ordered women (resp., men) that are all incomparable with another woman, then it is not possible to define a value $\alpha$ and an SMQ $Q$ such that the $\alpha$-stable marriages of $Q$ correspond to the weakly stable marriages of $P$. }

\subsection{Dominance and lex-male-optimality}
  
We recall that in SMPs a weakly-stable marriage dominates another 
weakly-stable marriage if men are happier (or equally happy) and there is at least a 
man that is strictly happier. The same holds for $\alpha$-stable marriages. 
As in SMPs there may be more than one undominated weakly-stable marriage, 
in SMQs there may be more than one undominated $\alpha$-stable marriage. 

%As in SMPs, an $\alpha$-stable marriage dominates another 
%$\alpha$-stable marriage men are happier (or equally happy) and there is at least a 
%man that is happier. Therefore, as for SMPs, there may be more than one undominated $\alpha$-stable %marriage. 

\begin{definition}[dominance]
Given two $\alpha$-stable marriages, 
say $M$ and $M'$, $M$ dominates $M'$ if every man 
is married in $M$ to more or equally preferred woman than in $M'$ 
and there is at least one man in $M$ married to a more preferred woman than in $M'$. 
\end{definition}

%If we consider $\alpha$-stability,  %rather than classical stability, 
%we may have {\em more than one undominated element} w.r.t. the classical dominance relation as shown in Example %\ref{ex-smw2}. However, the undominated 
%stable marriage obtained by removing the preference values from the problem 
%is contained in the set of the undominated $\alpha$-stable marriages of the problem with preference values. 

\begin{example}
\label{exalpha2}
Let us consider the SMQ shown in Example \ref{exalpha}. We recall that \red{$\alpha$ is $2$ and that}  the $\alpha$-stable marriages of this problem are  $M_1=\{(m_1,w_1),(m_2,w_2)\}$ and $M_2=\{(m_1,w_2),(m_2,w_1)\}$. 
%These marriages are undominated. 
%In fact, $M_1$ is not dominated by $M_2$ 
$M_2$ does not dominate $M_1$ since,  for $m_1$,  $M_1(m_1)>M_2(m_1)$ and 
%$M_2$ is not dominated by $M_1$ 
$M_1$ does not dominate $M_2$ since,  for $m_2$,  $M_2(m_2)>M_1(m_2)$. 
\hfill $\Box$ 
\end{example}

We now discriminate among the %undominated 
$\alpha$-stable marriages of an SMQ, 
by considering the preference values given by  women and men 
 to order pairs that differ for less than $\alpha$. %, i.e.,  pairs that are considered equally preferred.    

We will consider a marriage optimal %better than another one 
when the most popular men are as happy as possible and they are married with the most popular $\alpha$-feasible women.  

To compute a strict ordering on the men where the most popular men (resp., the most popular women) 
are ranked first, \red{we follow a reasoning similar to the one considered in \cite{aamas09,jaamas}, that is,}  we apply a voting rule \cite{handbook-sc} to the preferences given by the women (resp., by the men) . 
More precisely, such a voting rule takes in input the preference values given by the women 
over the men (resp., given by the men over the women) and returns a strict total order over the men (resp., women). 

\begin{definition}[lex-male-optimal] %w.r.t. $o_m$ and $o_w$]
\label{lex-male}
Consider an SMQ $P$, a natural number $\alpha$, and  
a voting rule $r$. 
Let us denote with $o_m$ (resp., $o_w$) the strict total order over the men (resp., over the women) computed by applying $r$ to the preference values that the women give to the men (resp., the men 
give to the women).  
%that orders first the most preferred men (resp., women) obtained by applying a voting rule over preference values that the women (resp., men) give to the men (resp., women).  
An $\alpha$-stable marriage $M$ is {\em lex-male-optimal} w.r.t. $o_m$ and $o_w$, 
%written as $M \succ_{lex_{(o_m,o_w)}}M$', 
if, for every other $\alpha$-stable marriage $M'$, the following conditions hold:  
%where $lex_{(o_m,o_w)}$ is the lexicographical order w.r.t. the orderings $o_m$ and $o_w$ defined as follows: 
\begin{itemize}
\item there is a man $m_i$ %, called the decisive man for $\succ_{lex_{(o_m,o_w)}}$, 
such that  $M(m_i)\succ_{o_{w}}M'(m_i)$, 
\item for every man $m_j \prec_{o_m} m_i$, $M(m_j)=M'(m_j)$. 
\end{itemize}
\end{definition}

\begin{proposition}
Given an SMQ $P$, a strict total ordering $o_m$ (resp., $o_w$) over the men (resp., women), 
\begin{itemize}
\item there is a unique lex-male-optimal $\alpha$-stable marriage w.r.t. $o_m$ and $o_w$, say $L$.
\item $L$ may be different from the male-optimal stable marriage of $c(P)$; 
\item if $\alpha(P)$ has a unique undominated weakly stable marriage, say $L'$, 
then $L$ coincides with $L'$, otherwise $L$ is one of the undominated weakly stable marriages of $\alpha(P)$. 
\end{itemize}
\end{proposition}

\begin{example}
Let us consider the SMQ, $P$, shown in Example \ref{exalpha}.
%\begin{itemize}
%\item $m_1:w_1 (3) > w_2 (2)$ 
%\item $m_2:w_1 (4) > w_2 (2)$,
%\item $w_1:m_2 (4) > m_1 (20)$,
%\item $w_2:m_1 (3) > m_2 (1).$
We have shown previously that this problem has two $\alpha$-weakly stable marriages that are undominated. 
We now want to discriminate among them by considering the lex-male-optimality notion. 
Let us consider as voting rule the rule that takes in input the preference values given by the women over the men (resp., by the men over the women) and returns a strict preference ordering over the men (resp., women). 
This preference ordering is induced by the overall score that each man (resp., woman) receives: 
men (women) that receive higher overall scores are more preferred. 
The overall score of a man $m$ (resp., woman $w$), say $s(m)$ (resp., $s(w)$),  
is computed by summing all the preference values that the women give to him (the men give to her). 
If two candidates receive the same overall score, we use a tie-breaking rule to order them.  
If we apply this voting rule to the preference values given by the women in $P$, then we obtain 
$s(m_1)=8+3=11$, $s(m_2)=5+1=6$, and thus the ordering $o_m$ is such that $m_1\succ_{o_m} m_2$.
 If we apply the same voting rule to the preference values given by the men in $P$, 
 $s(w_1)=3+4=7$, $s(w_2)=2+2=4$, and thus the ordering $o_w$ is such that $w_1\succ_{o_w} w_2$. 
The  lex-male-optimal $\alpha$-stable marriage w.r.t. $o_m$ and $o_w$ is the marriage $M_1=\{(m_1,w_1),(m_2,w_2)\}$.  
\hfill $\Box$
\end{example}

\subsection{Finding the lex-male-optimal $\alpha$-stable marriage}

It is possible to find  optimal $\alpha$-stable marriages by adapting the GS-algorithm for classical stable marriage problems \cite{gs}.

Given an SMQ $P$ and a natural number  $\alpha$, %where only the women gives preference values over the men, 
by Theorem \ref{teo-pa}, to find an $\alpha$-stable marriage it is sufficient to find a weakly stable marriage 
of $\alpha(P)$. This can be done by applying the GS algorithm to any linearization of $\alpha(P)$. 

%We can also adapt classical GS algorithm as follows: 
%\begin{itemize}
%\item the input is an SMQ, and 
%\item when a woman, say $w$, is married to a man, say $m$,
% and she receives a proposal by another man, say $m'$, 
% she rejects her current partner, i.e., $m$, if she prefers $m'$ more than $m$ 
% of at least degree $\alpha$, i.e., if $p(w,m')-p(w,m)\geq \alpha$. 
% \end{itemize}
% 
%We will call the algorithm above $\alpha$-GS algorithm. 
%
%We will now show that the marriage returned by $\alpha$-GS algorithm is $\alpha$-stable 
%and male-optimal, i.e., it is undominated by other $\alpha$-stable marriages  
%w.r.t. the classical dominance relation. 
%

Given an SMQ $P$, a natural number $\alpha$, and two orderings $o_m$ and $o_w$ over  men and  women computed by applying a voting rule to $P$ as described in Definition \ref{lex-male}, it is possible to find the $\alpha$-stable marriage that is lex-male-optimal w.r.t $o_m$ and $o_w$  by applying the GS algorithm to the linearization of $\alpha(P)$ where we order incomparable pairs, i.e., the pairs that differ for less than $\alpha$ in $P$, in accordance with the orderings $o_m$ and $o_w$. 

\begin{algorithm}[h*]
  \caption{{\em Lex-male-$\alpha$-stable-GS}\label{male-alg}}
%{\tiny %\small
\KwSty{Input}: 
$P$: an SMQ, 
$\alpha$: a natural number, 
$r$: a voting rule\\
\KwSty{Output}: $\mu$: a marriage\\
$o_m\gets$ the strict total order over the men obtained by applying $r$ to the preference values given by the women over the men\\
$o_w\gets$: the strict total order over the women obtained by applying $r$ to the preference values given by the men over the women\\
%$\alpha(P)\gets$ the SMP obatined from $P$ by putting incomparable the pairs that differ for less than $\alpha$\;
%Compute the ordering $o_m$ and $o_w$\;
$P'\gets$ the linearization of $\alpha(P)$ obtained by ordering incomparable pairs of $\alpha(P)$ 
in accordance with $o_m$ and $o_w$\;
$\mu \gets$ the marriage obtained by applying the GS algorithm to $P'$\;
\Return{$\mu$}
\end{algorithm}

\begin{proposition}
Given an SMQ $P$, a natural number $\alpha$, $o_m$ (resp., $o_w$) an ordering over the men (resp., women), 
algorithm {\em Lex-male-$\alpha$-stable-GS}  returns the lex-male-optimal 
$\alpha$-stable marriage of $P$ w.r.t. $o_m$ and $o_w$. 
\end{proposition}

%\proof{
%...........\hfill $\Box$\\}

%\section{Link-stability}
\section{Stability notions relying on links}
\label{sec-link}
\red{
Until now we have generalized the classical notion of stability by considering 
separately the preferences of the men and the preferences of the women. 
We now intend to define new notions of stability that 
% that does not generalize the classical notion of stability. 
%Such notions 
take into account simultaneously the preferences of the men and the women. 
Such a new notion will depend on 
the strength of the link of the married people, i.e.,  
how much a man and a woman want to be married with each other. 
This is useful to obtain a %`fairer' 
 new notion of stable marriage, that looks 
at the happiness of the pairs (man,woman) rather than at the happiness of the members of a single sex. 
}

%If we don't allow people to express also quantitative preferences, we are compelled to forget these preferences and %thus we may have a loss of information as shown in the following example.  

%\begin{example}
%\label{exw1}
%Let us consider an instance of SM with two men, say $m_1$ and $m_2$, and two women, say $w_1$ and $w_2$, 
%where men $m_1$ and $m_2$ give the same preference ranking over the women, say $o$. 
%Assume that $m_1$ and $m_2$ have different links with the women.  
%For example, $m_1$ has a very strong link with the first woman and 
%a very weak link with all the other women (i.e., $m_1$ prefers much more 
% to be married with the first woman in $o$ than with the other woman),  
% while $m_2$ has a link of similar strength for all the women 
% (i.e., he prefers to be married with the first woman in $o$ a little more than 
% with the second woman in $o$, %and so on, 
% but this preference ranking is not very strong for him). 
%%In classical SMs the two situations above are dealt in the same way, thus forgetting 
%%the additional information that $m_1$ and $m_2$ have in mind regarding the various women and thus the optimal %solutions returned by the classical algorithms do not reflect what people really desire. 
%We want instead to take into account these links to define a new stability notion. 
%\hfill $\Box$
%\end{example}
 
%To avoid this loss of information, we will define stable marriage problems 
%where people can also express quantitative preferences and we will consider such quantitative 
%preferences to define new notions of stability and optimality. 

A way to define the strength of the link of two people is the following. 

\begin{definition}[link additive-strength]
Given a man $m$ and a woman $w$, the {\em link additive-strength} of the pair $(m,w)$, denoted by $la(m,w)$,  is the value obtained by summing the preference value that $m$ gives to $w$
 and the preference value that $w$ gives to $m$, i.e., 
$la(m,w)= p(m,w)+ p(w,m)$.  
Given a marriage $M$, the {\em additive-link} of $M$, denoted by $la(M)$, is the sum of the links of all its pairs, i.e., $\sum_{\{(m,w)\in M\}}la(m,w)$. 
\end{definition}

Notice that we can use other operators beside the sum to define the link strength, \red{such as, for example, the maximum or the product. }%, and we can use various criteria to order the pairs on the basis of their links. } 
%3-3, 4-2: a parita' di somma e' meglio scegliere quello con il prodotto piu' alto. 
%La somma e il max sceglierebbe il secondo, mentre il prodotto sceglierebbe il primo, 
%perche' il prodotto, a parita' di somma, privilegia ler coppie con i numeri piu' vicini.

%Moreover, while now we consider the pairs with highest links as the best ones, we could also consider other optimality criteria. For example, we may be interested into the  equality of the preference values in the links. In such a case, we may consider a pair $(m,w)$ with equal preference values  $p(m,w)=p(w,m)=3$ better than a pair  $(m',w')$ with different preference values $p(m,w)=2$ and $p(w,m)=6$ even if the sum and the product of the preference values of $(m,w)$ is smaller than the ones of $(m',w')$.}

We now give a notion of stability that exploit the definition of the link additive-strength given above.

\begin{definition}[link-additive-stability]
Given a marriage $M$, a man $m$, and a woman $w$, 
the pair $(m,w)$ is a link-additive-blocking pair for $M$ if the following conditions hold: 
\begin{itemize}
\item $la(m,w)>la(m',w)$, 
\item $la(m,w)>la(m,w')$, 
\end{itemize}
where $m'$ is the partner of $w$ in $M$ and $w'$ is the partner of $m$ in $M$. 
A marriage is {\em link-additive-stable} if it does not contain link-additive-blocking pairs. 
\end{definition}

%\begin{example}
%Let $W=\{w_1,w_2,w_3\}$ and  $M=\{m_1,m_2,m_3\}$ 
%be respectively the set of women and men.   
%Consider the following instance of an SMQ: 
%\begin{itemize}
%\item $m_1:w_1 (9)  > w_2 (6)> w_3 (0.5)$,  
%\item $m_2:w_1 (2)  > w_3 (1)> w_2 (0.5) $,
%\item $m_3:w_2 (3)  > w_3 (2) > w_1 (0.5) $,
%\item $w_1:m_2 (3) > m_1 (1)> m_3 (0.5)$, 
%\item $w_2:m_1 (5) > m_2 (4)> m_3 (3).$
%\item $w_3:m_3 (9) > m_2 (8)> m_1 (1).$
%\end{itemize}
%A link2-stable marriage is 
%$M_1=\{(m_1,w_2), (m_2,w_1), (m_3,w_3)\}$. 
%The link of $M_1$ is $11+ 5+ 11=27$. 
%This is also the male-optimal stable marriage that we have when we don't consider the 
%preference values in the given problem. However, in general there are link2-stable marriages 
%that are not male-optimal for the problem without preference values as shown in the following example.
%\end{example}

\begin{example}
Let $\{w_1,w_2\}$ and  $\{m_1,m_2\}$ 
be, respectively, the set of women and men.   
Consider the following instance of an SMQ, $P$:  
\begin{itemize}
\item \red{$m_1:w_1^{[30]} >w_2^{[3]}$,  
\item $m_2:w_1^{[4]} >w_2^{[3]}$, %(2)$,
\item $w_1:m_2^{[6]} > m_1^{[5]}$, 
\item $w_2:m_1^{[10]} >m_2^{[2]}.$}
\end{itemize}
In this example there is a unique link-additive-stable marriage, that is 
$M_1=\{(m_1,w_1), (m_2,w_2)\}$, which has additive-link $la(M_1)=35+ 5=40$. 
%The marriage $M_2=\{(m_1,w_2), (m_2,w_1)\}$ is not link2-stable 
%since the pair $(m_1,w_2)$ is a link2-blocking pair. 
Notice that such a marriage has an additive-link higher than the male-optimal stable marriage of $c(P)$ that is 
$M_2=\{(m_1,w_2), (m_2,w_1)\}$ which has additive-link $la(M_2)=13+10=23$. \hfill $\Box$
%Therefore, if we want to maximize the happiness of the pairs (and not of a single sex), 
%and thus if we want to be `fairer',  it is reasonable to consider the new definition of stability that %relies 
%on the strength of the links of the pairs. \hfill $\Box$
% The new definition of stability allows us to obtain 'fairer' results than the classical male-optimal or
% female-optimal solutions that we can find if we don't consider the preference values.  
\end{example}

The strength of the link of a pair (man,woman), and thus the notion of link stability, can be also defined by considering the maximum operator instead of the sum operator. 

\begin{definition}[link maximal-strength]
\label{link2}
Given a man $m$ and a woman $w$, the {\em link maximal-strength} of the pair $(m,w)$, denoted by $lm(m,w)$,  is the value obtained by taking the maximum between the preference value that $m$ gives to $w$
 and the preference value that $w$ gives to $m$, i.e., 
$lm(m,w)=max( p(m,w), p(w,m))$.  
Given a marriage $M$, the {\em maximal-link} of $M$, denoted by $lm(M)$, is the maximum of the links of all its pairs, i.e., 
 $max_{\{(m,w)\in M\}}lm(m,w)$. 
\end{definition}

\begin{definition}[link-maximal-stability] 
Given a marriage $M$, a man $m$, and a woman $w$, 
the pair $(m,w)$ is a link-maximal-blocking pair for $M$ if the following conditions hold: 
\begin{itemize}
\item $lm(m,w)>lm(m',w)$, 
\item $lm(m,w)>lm(m,w')$, 
\end{itemize}
where $m'$ is the partner of $w$ in $M$ and $w'$ is the partner of $m$ in $M$. 
A marriage is {\em link-maximal-stable} if it does not contain link-maximal-blocking pairs. 
\end{definition}

%There are cases where there is a unique link-stable marriage. 
%If a stable marriage is unique, then it has also the highest link.
%
%\begin{example}
%Let $W=\{w_1,w_2,w_3\}$ and  $M=\{m_1,m_2,m_3\}$ 
%be respectively the set of women and men.   
%Consider the following instance of an SMQ: 
%\begin{itemize}
%\item $m_1:w_1 (9)  > w_2 (7)> w_3 (0.5)$,  
%\item $m_2:w_1 (2)  > w_3 (1)> w_2 (0.5) $,
%\item $m_3:w_2 (3)  > w_3 (2) > w_1 (0.5) $,
%\item $w_1:m_2 (7) > m_1 (1)> m_3 (0.5)$, 
%\item $w_2:m_1 (5) > m_2 (4)> m_3 (3).$
%\item $w_3:m_3 (9) > m_2 (7)> m_1 (1).$
%\end{itemize}
%The unique link-stable marriage of this problem is 
%$M_1=\{(m_1,w_2), (m_2,w_1), (m_3,w_3)\}$. 
%The link of $M_1$ is $12+ 9+ 11=32$. \hfill $\Box$
%\end{example}

\subsection{Relations with other stability notions}

Given an SMQ $P$, let us denote with $Linka(P)$ (resp., $Linkm(P)$) the stable marriage problem with ties obtained from $P$ by changing every preference value that a person $x$ gives to a person $y$ with the value 
$la(x,y)$ (resp., $lm(x,y)$), by changing the preference rankings accordingly, 
and by considering only these new preference rankings. 
%preference orderings induced by the new preference values.  

Let us denote with $I_{la}(P)$ (resp., $I_{lm}(P)$) the set of the link-additive-stable marriages 
(resp., link-maximal-stable marriages) of $P$ and 
with $I_w(Linka(P))$ (resp., $I_w(Linkm(P))$) the set of the weakly stable marriages of $Linka(P)$ (resp., $Linkm(P)$).  %where we don't consider the preference values. 
It is possible to show that these two sets coincide. 

\begin{theorem}
\label{link-w}
Given an SMQ $P$, $I_{la}(P)=I_w(Linka(P))$ and $I_{lm}(P)=I_w(Linkm(P))$.
\end{theorem}

\proof{Let us consider a marriage $M$. 
We first show that if $M\in I_w(Linka(P))$ then $M\in I_{la}(P)$. 
If $M\not\in I_{la}(P)$, there is a pair $(m,w)$ that is a link-additive-blocking pair, i.e., 
$la(m,w)> la(m,w')$ and $la(m,w)>la(m',w)$, where $w'$ (resp., $m'$) 
is the partner of $m$ (resp., $w$) in $M$. 
Since $la(m,w)> la(m,w')$, $m$ prefers $w$ to $w'$ in the problem $Linka(P)$,  
and, since $la(m,w)> la(m',w)$, $w$ prefers $m$ to $m'$ in the problem $Linka(P)$. 
Hence $(m,w)$ is a %classical 
blocking pair for the problem $Linka(P)$. %where we don't consider the preference values. 
Therefore, $M\not\in I_w(Linka(P))$. 

We now show that if $M\in I_{la}(P)$ then $M\in I_w(Linka(P))$. 
If $M \not \in I_w(Linka(P))$, there is a pair $(m,w)$ that is a blocking pair for 
$I_w(Linka(P))$, i.e.,  $m$ prefers $w$ to $w'$ in the problem $Linka(P)$, and 
$w$ prefers $m$ to $m'$ in the problem $Linka(P)$. By definition of the problem $Linka(P)$, 
$la(m,w)> la(m,w')$ and $la(m,w)>la(m',w)$. Therefore, $(m,w)$ is a link-additive-blocking pair for the problem $P$. 
Hence, $M\not\in I_{la}(P)$. 

It is possible to show similarly that $I_{lm}(P)=I_w(Linkm(P))$. \hfill $\Box$ \\
}

When no preference ordering changes in $Linka(P)$ (resp., $Linkm(P)$) w.r.t. $P$, 
then the link-additive-stable (resp., link-maximal-stable) marriages of $P$ 
coincide with the stable marriages of $c(P)$. %where we don't consider preference values. 

\begin{proposition}
Given an SMQ $P$, 
%when preference ranking of every man and every woman in $Linka(P)$  (resp., $Linkm(P)$)
%coincides with the one given in $P$, 
if $Linka(P)=c(P)$ ($Linkm(P)=c(P)$) , then $I_{la}(P)=I(c(P))$ (resp., $I_{lm}(P)=I(c(P))$). 
% and thus 
%there is a unique link-additive-stable (resp., link-maximal-stable) marriage for $P$ with the highest link (resp., link2),
% that  coincides with %the male-optimal stable marriage of $Linka(P)$, that coincides with 
%the male optimal of $P$ where we don't consider preference values.  
\end{proposition}

If there are no ties in $Linka(P)$ (resp., $Linkm(P)$), 
then there is a unique link-additive-stable marriage (resp., link-maximal-stable marriage) 
with the highest link.

\begin{proposition}
Given an SMQ $P$, if $Linka(P)$ (resp., $Linkm(P)$) has no ties, then there is a unique link-additive-stable 
(resp., link-maximal-stable) marriage with the highest link. 
\end{proposition}

%\proof{
%.............. \hfill $\Box$\\
%}

If we consider the definition of  link-maximal-stability, it is possible to define a class of SMQs where there is a  unique link-maximal-stable marriage with the highest link.

\begin{proposition}
In an SMQ $P$ where the preference values are all different, there is a unique link-maximal-stable marriage with the highest link. 
\end{proposition}

%\proof{
%...........\\
%}

%For every notion of stability that we have defined, 
%as for the classical SMs, a stable marriage is {\em male optimal} iff it is undominated by some other stable marriage. 
 
%every man %(resp., woman) 
%is paired with his %(resp., her) 
%highest ranked feasible partner, where the feasibility is based on the new notion of stability. 

%\section{Finding stable marriages}

%We will show that it is possible to solve stable marriage problems with quantitaive preferences, that are more complex
% than the classical stable marriages,  by adapting the well-known GS algorithm used to solve classical stable marriage %problems.  
%In particular, what will be changed w.r.t the classical GS algorithm will be the input and the choice strategy used %from a married woman when receives a proposal from another man. Such a strategy will depend 
%on the notion of stability and optimality that we are looking for.  
%Notice, that, while the classical GS algorithm %and its generalization in the context of partial orders 
%is not dependent from the order of the men that propose, our algorithms depend on such an order. 
%To have a fairer marriage, we will compute such an ordering on the basis of the preference values given by the women.
%Therefore, the highest man in such an ordering will be the most preferred by the women according to a certain voting %rule that considers their preference values. 
%
%
\subsection{Finding link-additive-stable and link-maximal-stable marriages with the highest link}

We now show that \red{ for some classes of preferences} it is possible to find optimal link-additive-stable marriages and link-maximal-stable marriages of an SMQ  by adapting algorithm GS,
which is usually used to find the male-optimal stable marriage in classical stable marriage problems.  

By Proposition \ref{link-w}, we know that the set of the link-additive-stable (resp., link-maximal-stable) marriages of an SMQ $P$ coincides with the set of the weakly stable marriages of the SMP $Linka(P)$ (resp., $Linkm(P)$). 
Therefore, to find a link-additive-stable (resp., link-maximal-stable) marriage, we can simply  apply algorithm GS to a linearization of $Linka(P)$ (resp., $Linkm(P)$).

\begin{algorithm}[h*]
  \caption{{\em link-additive-stable-GS (resp., link-maximal-stable-GS)}\label{link-alg}}
%{\tiny %\small
\KwSty{Input}: $P$: an SMQ\\
\KwSty{Output}: $\mu$: a marriage\\
$P'\gets Linka(P)$ (resp., $Linkm(P)$)\;
% the SMP obtained from $P$ by changing the preference value that a person $x$ gives to a person $y$ with the value 
%$Linka(x,y)$, by changing the preference ranking accordingly, and by forgetting the links' values \\
$P'' \gets$ a linearization of $P'$\; 
$\mu\gets$ the marriage obtained by applying GS algorithm to $P''$\;
\Return{$\mu$}
\end{algorithm}

\begin{proposition} 
Given an SMQ $P$, the marriage returned by algorithm {\em link-additive-stable-GS} ({\em link-maximal-stable-GS})  over $P$, say $M$,  is link-additive-stable (resp., link-maximal-stable).  %and male-optimal.  
%w.r.t. the profile with links. 
Moreover, if there are not ties in $Linka(P)$ (resp., $Linkm(P)$), 
$M$ is link-additive-stable (resp., link-maximal-stable) and it has the highest link.   
%If there are ties in $Linka(P)$, the returned marriage may have not the highest link.
\end{proposition} 

%\proof{..............\hfill $\Box$ \\}

\Omit{
%========================= omit=======================================================
..................Since $I_la(P)=I_c(Linka(P))$,  when there are no ties in $Linka(P)$, then 
every stable  solution of $Linka(P)$ is a link-additive-stable solution, and 
where there are ties in $Linka(P)$ every weakly stable solution of $Linka(P)$ is a link-additive-stable solution. 
If there are not ties in $Linka(P)$, the algorithm applies GS algorithm on $Linka(P)$, and, by definition of GS,  
the returned marriage is stable and male-optimal for $Linka(P)$, i.e., 
for every other $m$, for every other feasible marriage $M'$ of $Linka(P)$, 
$M(m)\geq M'(m)$ in $Linka(P)$, i.e., by definition of $Linka(P)$, $la(m,M(m))\geq la(m,M'm)$. 
Therefore, if we make the sum of every man $m$, 
we have that for every feasible marriage of $Linka(P)$, 
i.e., for every feasible marriage of $P$ (by Proposition \ref{link-w}),  
$la(M)\geq la(M')$ in $P$. Hence, if there are not ties in $Linka(P)$ link-additive-stable-GS returns the link-additive-stable marriage with the highest link. 

The results regarding ({\em link-maximal-stable-GS}) algorithm can be shown similarly. 
\hfill $\Box$\\ 
%If there are ties in $Linka(P)$ ......
}
%=====================================================================================

When there are no ties in $Linka(P)$ (resp., $Linkm(P)$), 
the marriage returned by algorithm {\em link-additive-stable-GS} (resp.,  {\em link-maximal-stable-GS}) is male-optimal w.r.t. the profile with links. \red{Such a marriage %a `fairer' marriage than 
may be different} from the classical male-optimal stable marriage of $c(P)$, 
%solution obatined without considering 
%where we don't consider links, 
since it considers the happiness of the men reordered 
according to their links with the women, rather than according  their single preferences. 

This holds, for example, when we assume to have an SMQ with preference values 
that are all different and we consider the notion of link2-stability. % as shown in the following proposition. 

\begin{proposition} 
Given an SMQ $P$ where the preference values are all different, 
the marriage returned by algorithm {\em link-maximal-stable-GS} algorithm over $P$ 
is link-maximal-stable and it has the highest link. 
\end{proposition} 

%\proof{ 
%...........................................\\}

%\begin{proposition}
%lex-link-male-optimality 
%\end{proposition}

%************* lex-male-optimal for link-stable solutions \\\\

%\subsection{Finding the lex-male-optimal link-stable (link2-stable) marriage}
%...............

\section{Conclusions and future work}
\label{sec-concl}
In this paper we have considered stable marriage problems with quantitative preferences, 
where both men and women can express a score over the members of the other sex. 
In particular, we have introduced new stability and optimality notions for such problems and we have compared them 
with the classical ones for stable marriage problems with totally or 
partially ordered preferences. Also, we have provided algorithms to find marriages that are optimal and 
stable according to these new %optimality and stability 
notions by adapting the Gale-Shapley algorithm.

We have also considered an optimality notion 
(that is, lex-male-optimality)
that exploits a voting rule to linearize the partial
orders. We intend to study if this use of voting rules within
stable marriage problems may have other benefits. 
In particular, we want to investigate if
the procedure defined to find such an optimality notion 
inherits the properties of the voting rule with respect to manipulation:
we intend to check whether, if the voting rule is NP-hard to manipulate,
then also the procedure on SMQ that exploits such a rule is  NP-hard to manipulate. 
This would allow us to transfer several existing results on 
manipulation complexity, which have been obtained for voting rules, 
to the context of procedures to solve stable marriage problems with quantitative preferences.

%We plan to identify other stability and optimality notions and to characterize them in terms of the classical ones. 
%Moreover, we plan to investigate manipulat
%In this paper we have defined an  optimality notion that depends on a voting rule and we have defined a procedure to find a marriage that is optimal according to such a notion. 
%We plan to investigate the relation between the manipulability of this procedure and the manipulability of the voting rule.  

\section*{Acnowledgements}

This work has been partially supported by the MIUR PRIN 20089M932N project 
``Innovative and multi-disciplinary approaches for constraint and
preference reasoning''.

\bibliography{bib-stable1} 

%\begin{contact}
%Maria Silvia Pini, Francesca Rossi, K. Brent Venable\\
%Department of Pure and Applied Mathematics\\
%University of Padova, Italy\\
%\email{{mpini,frossi,kvenable}@math.unipd.it}
%\end{contact}

\begin{contact}
Maria Silvia Pini\\
Department of Pure and Applied Mathematics\\
University of Padova, Italy\\
\email{mpini@math.unipd.it}
\end{contact}

\begin{contact}
Francesca Rossi\\
Department of Pure and Applied Mathematics\\
University of Padova, Italy\\
\email{frossi@math.unipd.it}
\end{contact}

\begin{contact}
K. Brent Venable\\
Department of Pure and Applied Mathematics\\
University of Padova, Italy\\
\email{kvenable@math.unipd.it}
\end{contact}

\begin{contact}
Toby Walsh\\
NICTA and UNSW, Sydney, Australia\\ 
\email{Toby.Walsh@nicta.com.au}
\end{contact}

\end{document}